\documentclass[10pt,twocolumn,letterpaper]{article}

\usepackage{cvpr}
\usepackage{times}
\usepackage{epsfig}
\usepackage{graphicx}
\usepackage{amsmath}
\usepackage{amssymb}

\usepackage[caption=false,font=scriptsize]{subfig}
\usepackage[linesnumbered,lined,boxed,commentsnumbered]{algorithm2e}

\DeclareMathOperator*{\argmin}{arg\,min}

\cvprfinalcopy 

\def\cvprPaperID{903} 

\SetAlCapSkip{0.1cm}
\begin{document}

\title{Dynamic Attention-controlled Cascaded Shape Regression Exploiting Training Data Augmentation and Fuzzy-set Sample Weighting}

\author{Zhen-Hua Feng$^1$  ~~Josef Kittler$^1$  ~~William Christmas$^1$  ~~Patrik Huber$^1$  ~~Xiao-Jun Wu$^2$\\
$^1$ Centre for Vision, Speech and Signal Processing, University of Surrey, Guildford GU2 7XH, UK\\
$^2$School of IoT Engineering, Jiangnan University, Wuxi 214122, China\\
{\tt\small \{z.feng, j.kittler, w.christmas, p.huber\}@surrey.ac.uk, wu\_xiaojun@jiangnan.edu.cn}
}

\maketitle

\begin{abstract}
We present a new Cascaded Shape Regression (CSR) architecture, namely Dynamic Attention-Controlled CSR (DAC-CSR), for robust facial landmark detection on unconstrained faces. Our DAC-CSR divides facial landmark detection into three cascaded sub-tasks: face bounding box refinement, general CSR and attention-controlled CSR. The first two stages refine initial face bounding boxes and output intermediate facial landmarks. Then, an online dynamic model selection method is used to choose appropriate domain-specific CSRs for further landmark refinement. The key innovation of our DAC-CSR is the fault-tolerant mechanism, using fuzzy set sample weighting, for attention-controlled domain-specific model training. Moreover, we advocate data augmentation with a simple but effective 2D profile face generator, and context-aware feature extraction for better facial feature representation. Experimental results obtained on challenging datasets demonstrate the merits of our DAC-CSR over the state-of-the-art methods.
\end{abstract}

\section{Introduction}
Facial Landmark Detection (FLD), also known as face alignment, is a prerequisite for many automatic face analysis systems, \eg face recognition~\cite{beveridge2015report,Sun_2014_CVPR,Taigman2014}, expression analysis~\cite{eleftheriadis_accv,eleftheriadis2015discriminative} and 2D-3D inverse rendering~\cite{Aldrian2013,hu2017efficient,huber2015fitting,kittler20163d,Piotraschke_2016_CVPR,zhu2015discriminative}.
Facial landmarks provide accurate shape information with semantic meaning, enabling geometric image normalisation and feature extraction for use in the remaining stages of a face processing pipeline.
This is crucial for high-fidelity face image analysis.
As the technology of FLD for constrained faces has already been well developed, the current trend is to address FLD for unconstrained faces in the presence of extreme variations in pose, expression, illumination and partial occlusion~\cite{Belhumeur2011,Burgos-Artizzu2013,aflw_dataset_2011,le2012interactive,sagonas2016300}.
\begin{figure}[t]
\centering
   \includegraphics[trim={0mm 136mm 120mm 0mm}, clip, width=.9\linewidth]{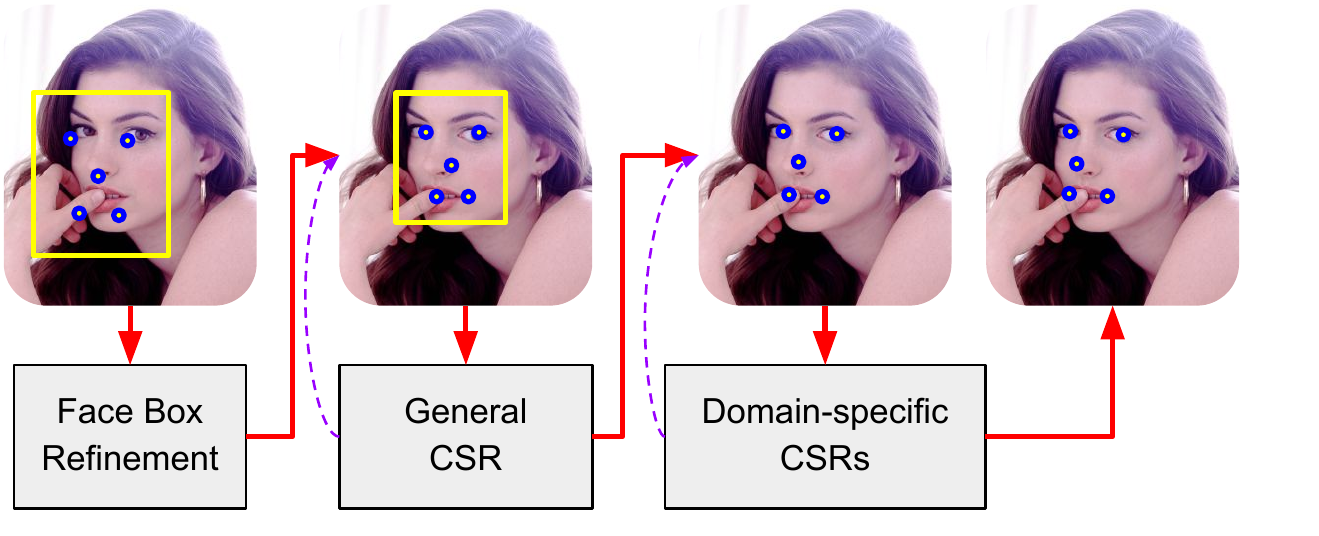}
   \caption{The pipeline of our proposed DAC-CSR.}
\label{fig_0}
\end{figure}

More recently, unconstrained FLD has seen huge progress owing to the state-of-the-art Cascaded Shape Regression (CSR) architecture~\cite{Cao2014face,dollar2010cascaded,feng2015cascaded,ren2014face,zhu2016unconstrained}. 
The key to the success of CSR is to construct a strong regressor from a set of weak regressors arranged in a cascade.
This architecture greatly improves the performance of FLD in terms of generalisation capacity and accuracy.
However, in the light of very recent studies~\cite{Trigeorgis_2016_CVPR,xiong2015global,Zhang_2016_CVPR,zhu2016unconstrained,zhu2016face}, the capacity of CSR appears to be saturating, especially for unconstrained faces with extreme appearance variations.
For example, the FLD error of state-of-the-art CSR-based methods increases from around 3\% (error in percent of the inter-ocular distance) on the Labelled Face Parts in the Wild (LFPW)~\cite{Belhumeur2011} dataset to 6.5\% on the more challenging Caltech Occluded Faces in the Wild (COFW)~\cite{Burgos-Artizzu2013} dataset.
This degradation has three main reasons:
1) The modelling capacity of the existing CSR architecture is limited.
2) CSR is sensitive to the positioning of face bounding boxes used for landmark initialisation.
3) The volume of available training data is insufficient.
Can these limitations be overcome, especially for unconstrained faces exhibiting extreme appearance variations? 
We offer an encouraging answer by presenting a new Dynamic Attention-Controlled CSR (DAC-CSR) architecture with a dynamic domain selection mechanism and a novel training strategy which benefits from training data augmentation and fuzzy set training sample weighting. 

Fig.~\ref{fig_0} depicts a simplified overview of the proposed DAC-CSR architecture.
Its innovation is in linking three types of regressor cascades performing in succession: 1) face bounding box refinement for
better landmark initialisation, 2) an 
initial landmark update using a general CSR,  and 3) a final landmark refinement by dynamically selecting an attention-controlled domain-specific
CSR that is optimised to improve landmark location estimates. The new architecture decomposes
the task at hand into three cascaded sub-tasks that are easier
to handle. 

In contrast to previous multi-view models, \eg~\cite{xiong2015global,zhu2016unconstrained}, the key innovation of our DAC-CSR is its in-built
fault-tolerant mechanism. The fault tolerance is achieved by means of an innovative training strategy  for attention-controlled model training of the set of domain-specific CSRs performing the final shape update refinement. Rather than using samples from just a single domain, each domain-specific regressor cascade is trained using all the training samples. However, their influence is controlled by a domain-specific fuzzy membership function which weighs samples from the relevant domain more heavily than all the other training samples. An annealing schedule of domain-specific fuzzy membership functions progressively sharpens the relative weighting of in-domain and out-of-domain training samples in favour of the in-domain set for successive stages of each domain-specific cascade. 

Each test sample progresses through the system of cascades. Prior to each of the domain-specific  cascade stages, the domain of attention is selected dynamically based on the current shape estimate.  The proposed training strategy guarantees that each domain-specific cascaded regressor can cope with out-of-domain test samples and is endowed with the capacity to update the shape in the correct direction even if the current domain has been selected subject to labelling error. This is the essence of error tolerance of the proposed system.

An important contributing factor to the promising performance of our DAC-CSR is training data augmentation. Our innovation here is to use a 2D face model for synthesising extreme profile face poses (out of plane rotation) with realistic background. Furthermore, we propose a novel context-aware feature extraction
method to extract rich local facial features in the context of global face description.

The proposed framework has been evaluated on benchmarking databases using standard protocols. The results achieved on the database containing images with extreme poses (AFLW~\cite{aflw_dataset_2011}) are significantly better than the state-of-the-art performance reported in the literature.

The paper is organised as follows. In the next section we present a brief review of related literature. The preliminaries of CSR are presented in Section~\ref{sec_csr}. The proposed DAC-CSR is introduced in Section~\ref{sec_4_1}. The discussion of its training is confined to Section~\ref{sec_fdcsr}, which defines the domain-specific fuzzy membership functions and their annealing schedule. On-line dynamic domain selection is the subject of Section~\ref{sec_domain_pre} and the proposed feature extraction scheme can be found in Section~\ref{sec_4_4}. Section~\ref{sec_5} addresses the problem of training set augmentation. The experimental evaluation carried out and the results achieved are described in Section~\ref{sec_6}. The paper is drawn to conclusion in Section~\ref{sec_7}. 

\section{Related Work}
\label{sec_related_work}
Facial landmark detection can trace its history to the nineteen nineties. The representative FLD methods making the early milestones include Active Shape Model (ASM)~\cite{cootes1992active}, Active Appearance Model (AAM)~\cite{cootes1998active} and Constrained Local Model (CLM)~\cite{Cristinacce2006}.
These algorithms and their extensions have achieved excellent FLD results in constrained scenarios~\cite{feng2012automatic}.
As a result, the current trend is to develop a more robust FLD for unconstrained faces that are rich in appearance variations.
The leading algorithms for unconstrained FLD are CSR-based approaches~\cite{Cao2014face,dollar2010cascaded,feng2015cascaded,ren2014face,zhu2016unconstrained}. 
In contrast to the classical methods such as ASM, AAM and CLM that rely on a generative PCA-based shape model, CSR directly positions facial landmarks on their optimal locations based on image features.
The shape update is achieved in a discriminative way by constructing a mapping function from robust shape-related local features to shape updates.
The secret of the success of CSR is the architecture that cascades a set of weak regressors in series to form a strong regressor. 

There have been a number of improvements to the performance of CSR-based FLD.
One category of these improvements is to enhance some components of the existing CSR architecture.
For example, the use of more robust shape-related local features, \eg Scale-Invariant Feature Transform (SIFT)~\cite{xiong2013supervised,Zhang_2016_CVPR,zhang2014coarse}, Histogram of Oriented Gradients (HOG)~\cite{deng2016m,feng2015cascaded,huber2015fitting,Yan2013}, Sparse Auto-Encoder (SAE)~\cite{FENG2015}, Local Binary Features (LBF)~\cite{Cao2014face,ren2014face} and Convolutional Neural Networks (CNN-) based features~\cite{Trigeorgis_2016_CVPR,Xiao2016}, has been suggested.
Another example is to use more powerful regression methods as weak regressors in CSR, such as random forests~\cite{Cao2014face,ren2014face} and deep neural networks~\cite{Sun2013,Trigeorgis_2016_CVPR,Xiao2016,Zhang_2016_CVPR,zhang2014coarse,zhang2014facial}.
Lately, 3D face models have been shown to positively impact FLD in challenging benchmarking datasets, especially in relation to faces with extreme poses~\cite{feng2015cascaded,liu2016joint,zhu2016face}.

\textbf{Multi-view models:}
Another important approach is to adopt advanced CSR architectures, such as the use of multiple CSR models or constructing multi-view models.
Feng \etal~\cite{FENG2015} constructed multiple CSR models by randomly selecting subsets from the original training set and fusing multiple outputs to produce the final FLD result.
A similar idea has also been used in~\cite{yang2015random}.
As an alternative, a multi-view FLD system employs a set of view-specific models that are able to achieve more accurate landmark detection for faces exhibiting specific
views~\cite{cootes2002view,Tuzel2016,zhu2016unconstrained}.

However, the use of multiple models or multi-view models is not without difficulties.
One has to either estimate the view of a test image to select an appropriate model, or apply all view-specific models to a test image and then choose the best result as the final output.
For the former, implementing a model selection stage for unconstrained faces is hard in practice.
An erroneously selected view-specific model may result in FLD failure.
For the latter strategy, it is time-consuming to apply all the trained models to a test image.
Also, the ranking of the outputs of different view-based models is non-trivial.
In contrast to previous studies, our DAC-CSR addresses these issues by improving the fault-tolerance properties of a trained domain-specific model and by using an online dynamic model selection strategy. 

\textbf{Data augmentation:}
For a learning-based approach such as CSR, the availability of a large volume of training samples is essential.
However, it is a tedious task to manually annotate facial landmarks for a large quantity of training data.
To address this problem, data augmentation is widely used in CSR-based FLD.
Traditional methods include random perturbation of initial landmarks, image flipping, image rotation, image blurring and adding noise to the original face images.
However, none of these methods are able to inject new out-of-plane rotated faces to an existing training dataset.
Recently, to augment a training set by samples with rich pose variations, the use of 3D face models has been suggested.
For instance, Feng \etal~\cite{feng2015cascaded,kittler20163d,song2016dictionary} used a 3D morphable face model to synthesise a large number of 2D faces.
However, the synthesised virtual faces lack realistic appearance variations especially in terms of background and expression changes.
To mitigate this problem, they advocated a cascaded collaborative regression strategy to train a CSR from a mixture of real and synthesised faces.
To generate realistic face images with pose variations, Zhu \etal fit a 3D shape model to 2D face images and generate profile face views from the reconstructed 3D shape information~\cite{zhu2016face}.
However, these 3D-based methods~\cite{feng2015cascaded,kittler20163d,zhu2016face} require 3D face scans for model construction, which are expensive to capture.
Also, it is difficult in practice to fit a 3D face model to 2D images.
In this paper, we propose a simple but efficient 2D-based method to generate virtual faces with out-of-plane pose variations.
\begin{figure*}[t]
\centering
   \includegraphics[trim={0mm 61mm 1mm 0mm}, clip, width=.9\linewidth]{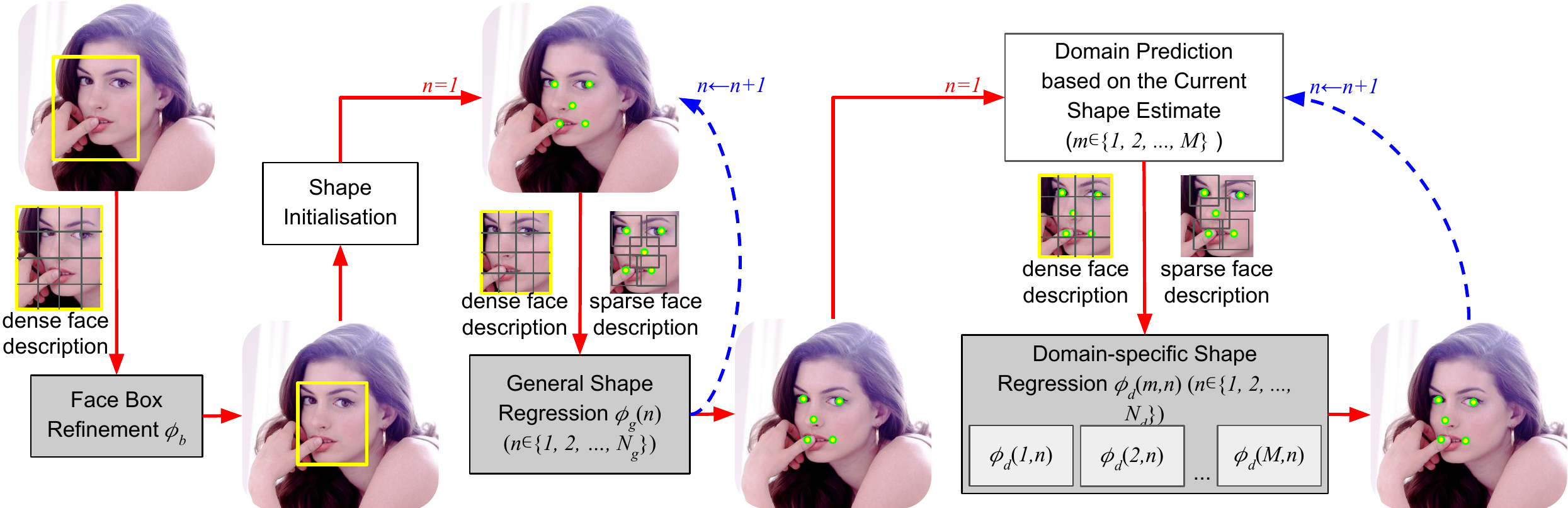}
   \caption{The proposed DAC-CSR has three stages in cascade: face bounding box refinement, general CSR and domain-specific CSR.}
\label{fig_1}
\end{figure*}

\section{Cascaded Shape Regression (CSR)}
\label{sec_csr}
Given an input face image $\mathbf{I}$ and the corresponding face bounding box $\mathbf{b} = [x_1, y_1, x_2, y_2]^T$ (coordinates of the upper left and lower right corners) of a detected face in the image, the goal of FLD is to output the face shape in the form of a vector, $\mathbf{s} = [x_1, y_1, ..., x_L, y_L]^T$, consisting of the coordinates of $L$ pre-defined facial landmarks with semantic meaning such as eye centres and nose tip.
To this end, we first initialise the face shape, $\mathbf{s}'$, by putting the mean shape into the bounding box.
Then a trained CSR $\Phi = \{\phi(1), \phi(2), ..., \phi(N)\}$ is used to update the initial shape estimate, where $\Phi$ is a strong regressor consisting of $N$ weak regressors.
A weak regressor can be obtained using any regression method, such as linear regression, random forests and neural networks.
In this paper, we use ridge regression as a weak regressor, \ie $\phi = \{\mathbf{A}, \mathbf{e}\}$:
\begin{equation}
\phi: \delta\mathbf{s} = \mathbf{A}\cdot f(\mathbf{I},\mathbf{s}') + \mathbf{e},
\end{equation}
where $\mathbf{A} \in \mathbb{R}^{2L \times N_f}$ is a projection matrix, $N_f$ is the dimensionality of a shape-related feature vector extracted using $f(\mathbf{I},\mathbf{s}')$, and $\mathbf{e} \in \mathbb{R}^{2L}$ is an offset.
For the shape-related feature extraction, we apply local descriptors, \eg HOG, to the neighbourhoods of all the facial landmarks of the current shape estimate and concatenate the extracted features into a long vector.
The use of a weak regressor results in an update to the current shape estimate:
\begin{equation}
\mathbf{s}' \leftarrow \mathbf{s}' + \delta \mathbf{s}.
\end{equation}
A trained CSR applies all the weak regressors in cascade to progressively update the shape estimate and obtain the final FLD result from an input image.

Given a training dataset $T = \{\mathbf{I}_i, \mathbf{b}_i, \mathbf{s}^*_i\}_{i=1}^{I}$ with $I$ samples including face images, face bounding boxes and manually annotated facial landmarks, we first obtain the initial shape estimates, $\{\mathbf{s}'_i\}_{i=1}^{I}$, of all the training samples using the face bounding boxes provided.
Then the shape update between the current shape estimate and ground-truth shape of the $i$th training sample can be calculated using $\delta \mathbf{s}^*_i = \mathbf{s}^*_i - \mathbf{s}'_i$.
The first weak regressor is obtained using ridge regression by minimising the loss:
\begin{equation}
\argmin_{\mathbf{A}, \mathbf{e}} \sum_{i = 1}^{I}
||\mathbf{A} \cdot f(\mathbf{I}_i, \mathbf{s}'_i) + \mathbf{e} - \delta\mathbf{s}^*_i ||^2_2 + \lambda ||\mathbf{A}||_F^2,
\label{eq:csr-loss}
\end{equation}
where $\lambda$ is the weight of the regularisation term.
This is a typical least-square estimation problem with a closed-form solution~\cite{FENG2015,xiong2013supervised}.
Last, this trained weak regressor is used to update the current shape estimates of all the training samples, which forms the training data for the second weak regressor.
This procedure is repeated until all the $N$ weak regressors are obtained.

\section{Dynamic Attention-controlled CSR}
\subsection{Architecture}
\label{sec_4_1}
The architecture of the proposed DAC-CSR method has three cascaded stages: face bounding box refinement, general CSR and domain-specific CSR, as shown in Fig.~\ref{fig_1}. 
In fact, our DAC-CSR can be portrayed as a strong regressor $\Phi = \{\phi_b, \Phi_g, \Phi_d\}$, where $\phi_b$ is a weak regressor for face bounding box refinement, $\Phi_g = \{ \phi_g(1), ..., \phi_g(N_g)\}$ is a classical CSR with $N_g$ weak regressors,
$\Phi_d = \{ \Phi_d(1), ..., \Phi_d(M) \}$ is a strong regressor with $M$ domain-specific CSRs and each of them has $N_d$ weak regressors $\Phi_d(m) = \{ \phi_d(m,1), ..., \phi_d(m,N_d) \}$.

\textbf{Face bounding box refinement:} We define the weak regressor for the first step as $\phi_b = \{\mathbf{A}_b, \mathbf{e}_b\}$:
\begin{equation}
\phi_b: \delta\mathbf{b} = \mathbf{A}_b \cdot f_{b}(\mathbf{I},\mathbf{b}) + \mathbf{e}_b,
\end{equation}
where $f_{b}(\mathbf{I},\mathbf{b})$ extracts dense local features from the image region inside the original face bounding box and $\delta \mathbf{b}$ is used to adjust the original bounding box.

The training of this weak regressor is the same as the procedure introduced in Section~\ref{sec_csr} for classical CSR.
The only difference here is that we use face bounding box differences instead of shape differences for the regressor learning in Eq.~(\ref{eq:csr-loss}).
The ground-truth face bounding box for a training sample is computed by taking the minimum enclosing rectangle around the ground-truth face shape.

\textbf{General CSR:}
The initial shape estimate, $\mathbf{s}'$, for general CSR is obtained by translating and scaling the mean shape so that it exactly fits into the refined bounding box, touching all four sides.
Then the general CSR progressively updates the initial shape estimate, $\mathbf{s}' \leftarrow \mathbf{s}' + \delta\mathbf{s}$ , using all the weak regressors in $\Phi_g = \{\phi_g(1), ..., \phi_g(N_g)\}$, as indicated in Algorithm~\ref{algorithm_1}.
The $n$th weak regressor is defined as $\phi_g(n) = \{\mathbf{A}_g(n), \mathbf{e}_g(n)\}$:
\begin{equation}
\phi_g(n): \delta\mathbf{s} = \mathbf{A}_g(n) \cdot f_{c}(\mathbf{I},\mathbf{s}') + \mathbf{e}_g(n),
\end{equation}
where $f_{c}(\mathbf{I},\mathbf{s}')$ is a context-aware feature extraction function that combines both dense face description and shape-related sparse local features.
The training of this stage is the same as the classical CSR introduced in Section~\ref{sec_csr}.
\begin{algorithm}[t]
\label{algorithm_1}
\caption{FLD using our DAC-CSR.}
\SetKwInOut{Input}{input}
\SetKwInOut{Output}{output}
 \Input{image $\mathbf{I}$, face bounding box $\mathbf{b}$ and a trained DAC-CSR model $\Phi = \{\phi_b, \Phi_g, \Phi_d\}$}
 \Output{facial landmarks $\mathbf{s}'$}
 refine the face bounding box $\mathbf{b}$ using $\phi_b$ \;
 estimate the current face shape, $\mathbf{s}'$, using the refined face bounding box \;
 \For{$n\leftarrow 1$ \KwTo $N_g$}{
 apply the $n$th general weak regressor $\phi_g(n)$ to update the current shape estimate\;
 }
 \For{$n\leftarrow 1$ \KwTo $N_d$}{
 predict the label ($m$) of the sub-domain of the current shape estimate using Eq.~(\ref{equ_label_prediction_2}) \;
  apply the $n$th weak regressor $\phi_d(m,n)$ in the $m$th domain-specific CSR to update the current shape\;
  }
\end{algorithm}

\textbf{Domain-specific CSR:}
Suppose this stage has $M$ domain-specific CSRs corresponding to $M$ sub-domains, each having $N_d$ weak regressors.
The $n$th weak regressor of the $m$th domain-specific CSR is defined as:
\begin{equation}
\phi_d(m,n): \delta\mathbf{s} = \mathbf{A}_d(m,n) \cdot f_{c}(\mathbf{I},\mathbf{s}') + \mathbf{e}_d(m,n),
\end{equation}
where $m = 1, ..., M$, $N = 1, ..., N_d$.
Given the current shape estimate $\mathbf{s}'$ output by the previous general CSR, a domain predictor is used to select a domain-specific CSR for the current shape update (Section~\ref{sec_domain_pre}).
It should be noted that we use a dynamic domain selection strategy, which updates the label for the domain-specific model selection after each shape update, as shown in Algorithm~\ref{algorithm_1}.
As a result of the proposed domain-specific CSR training described in Section~\ref{sec_fdcsr}, this mechanism makes our DAC-CSR tolerant to domain prediction errors.

\subsection{Offline Domain-specific CSR Training}
\label{sec_fdcsr}
\begin{figure}[t]
\centering
   \includegraphics[trim={0mm 80mm 165mm 0mm}, clip, width=.7\linewidth]{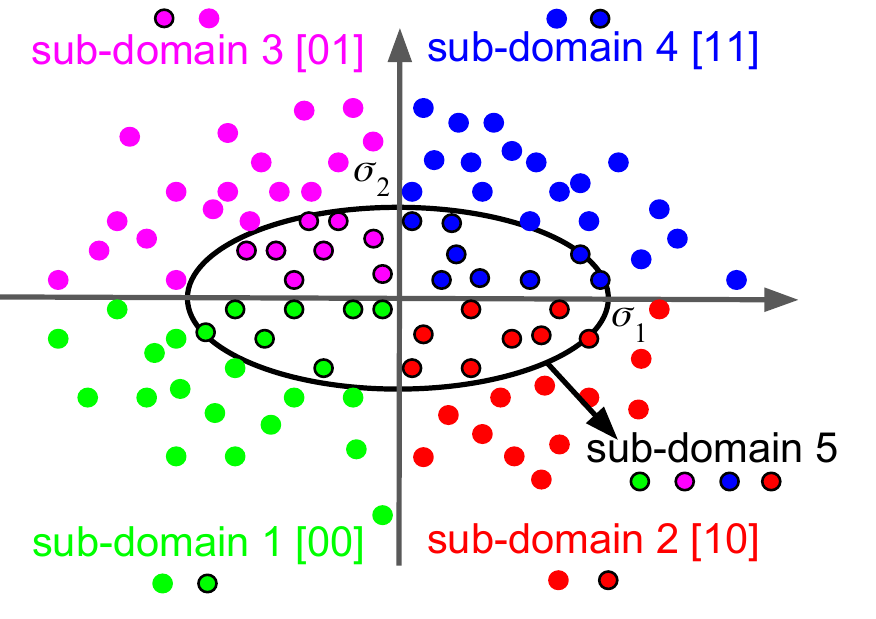}
   \caption{The proposed domain split strategy ($K = 2$, $\overline{c}_k=0$).}
\label{fig_2}
\end{figure}
Given a training dataset $T$ with $I$ samples, as introduced in Section~\ref{sec_csr}, the first two stages, \ie face bounding box refinement and general CSR, are trained directly using $T$.
To train a domain-specific CSR, we first create $M$ subsets $\{T_1, ..., T_{M}\}$ from the original training set, where $T_m \subseteq T$.
To this end, we normalise all the current shape estimates, output by the previous general CSR, to the interval $[0,1]$.
Then PCA is used to obtain the first $K$ shape eigenvectors.
All the current shape estimates are projected to the $K$-dimensional subspace to obtain the projected coefficients $\{\mathbf{c}_i\}_{i=1}^{I}$, where $\mathbf{c}_i = [c_{i,1}, ..., c_{i,K}]^T$.
Then the original domain is partitioned into $M=2^K+1$ overlapping sub-domains, as demonstrated in Fig.~\ref{fig_2} for $K=2$.
For the $M$th sub-domain, it includes the training samples satisfying $\sum_{k=1}^{K} \frac{(c_{i,k} - \overline{c}_{k})^2}{(\sigma(k))^2} \leq 1$, where $\overline{c}_{k}$ and $\sigma(k)$ are the mean and standard deviation of the $k$th element of the coefficient vectors.
For other sub-domains, each includes the training samples in a specific region of a K-dimensional coordinate system.
To be more specific, for each coefficient $\mathbf{c}_i$, a sub-domain membership word $g(\mathbf{c}_i)$ is generated by:
\begin{equation}
g(\mathbf{c}_i) = 1 + \sum_{k=1}^K b_c(c_{i,k})2^{k-1},
\end{equation}
where $b_c(c_{i,k})$ is a coding function that converts the $k$th element in a coefficient vector to a bit:
\begin{equation}
b_c(c_{i,k}) 
=
\left\{
\begin{array}{ll}
    1 & \text{if } c_{i,k} \geq  \overline{c}(k)\\
    0 & \text{otherwise}
\end{array} 
\right..
\end{equation}
Then the $m$th sub-domain, $1 < m < 2^K$, includes the training samples with their membership words $g(\mathbf{c}_i) = m$.
Our domain split strategy results in $M$ sub-domains with overlapping boundaries.
This is different from previous studies using multi-view models such as~\cite{xiong2015global,zhu2016unconstrained},
in which the intersection of any two different subsets is empty, \ie $T_i \cap T_j = \oslash, \forall i \neq j$.

The advantage of our domain split strategy is that it improves the fault-tolerance ability of each trained domain-attention model, because of the overlap of two different sub-domains.
For a test sample, a domain predictor may output an inaccurate label for model selection due to the rough shape estimate provided from the previous general CSR.
But, the inaccurately selected domain-specific model is still able to refine the current shape estimate.
To further improve this refinement capacity, we propose a fuzzy training strategy.
For each domain-specific CSR, we use all the training samples from the original training set to train a specific regressor, but weight more heavily the training samples of the specific domain by increasing their fuzzy set membership values in the objective function.
More specifically, to train the $n$th weak regressor of the $m$th domain-specific CSR, the objective function is defined as:
\begin{equation}
\argmin_{\mathbf{A_d}, \mathbf{e_d}} \sum_{i = 1}^{I}
w_i||\mathbf{A_d} \cdot f_{c}(\mathbf{I}_i, \mathbf{s}'_i) + \mathbf{e_d} - \delta\mathbf{s}^*_i ||^2_2 + \lambda ||\mathbf{A}_d||_F^2,
\end{equation}
where $w_i$ is a fuzzy set membership value defined by:
\begin{equation}
w_i = \left\{
\begin{array}{ll}
1-h(n)   & \text{if } \{\mathbf{I}_i, \mathbf{b}_i, \mathbf{s}^*_i\} \in T_m\\
h(n) & \text{otherwise}
\end{array}
\right.,
\end{equation}
where $h(n)$ is a decreasing function which progressively reduces the weights of the training samples not belonging to the $m$th sub-domain and increases the weights of the training samples of the $m$th sub-domain.
This is a standard weighted least-square estimation problem with a closed-form solution.
It should be noted that our fuzzy domain-specific model learns a weak regressor that is able to refine a face shape estimate from any sub-domain, and with better capacity to refine face shapes from a specific domain.
This capability is exhibited even when using a domain split strategy without overlap.

\subsection{Dynamic Domain Selection in Testing}
\label{sec_domain_pre}
Given a new test image with a detected face bounding box, the trained DAC-CSR model $\Phi = \{\phi_b, \Phi_g, \Phi_d\}$ first applies the face bounding box refiner $\phi_b$ and general CSR $\Phi_g$ to obtain the intermediate face shape estimate $\mathbf{s}'$.
Then a specific domain-attention weak regressor is selected to further update the current shape estimate.

To select an appropriate weak regressor, the current shape estimate $\mathbf{s}'$ is projected into the PCA space learned at training time to obtain the coefficient vector $\mathbf{c}$, and the label of the sub-domain is obtained using:
\begin{equation}
\label{equ_label_prediction_2}
p(\mathbf{c}) = 
\left\{
\begin{array}{ll}
    2^K + 1 & \text{if } \sum_{k=1}^{K} \frac{(c_{i,k} - \overline{c}_{k})^2}{(\sigma(k))^2} \leq 1\\
    g(\mathbf{c}) & \text{otherwise}
\end{array} 
\right..
\end{equation}
Note that, here, the sub-domains are not overlapped.
This is different from the domain split strategy used in the training stage.
However, this domain prediction function is only based on the current shape information and may provide inaccurate labels for model selection.
To address this issue and further improve the fault-tolerance capacity of our DAC-CSR, a dynamic domain selection strategy is used.

As discussed in the last section, a trained domain-specific CSR is able to improve the current shape estimate even if selected in error by the domain prediction mechanism.
Hence the updated shape estimate produced by the $n$th weak regressor can be a basis for selecting a more appropriate domain in the next step of the shape updating process.
We re-run the domain prediction before performing the next weak regressor and choose the $(n+1)$st weak regressor of a newly selected domain-specific model for current shape update, as summarised in Algorithm~\ref{algorithm_1}.
This dynamic model selection strategy is repeated after each shape update in our domain-specific CSR.

\subsection{Context-aware Feature Extraction}
\label{sec_4_4}
Feature extraction is crucial for constructing a robust mapping from feature space to shape updates.
In classical CSR-based approaches, shape-related local features are created by concatenating all the extracted local features around each landmark into a long vector.
Although this sparse shape-related feature extraction method provides a good description of the texture information of different facial parts, it does not offer a good representation of the contextual information of faces.
In our DAC-CSR, we use a context-aware feature extraction method.
To be more specific, we use both a dense local description of the whole face region and sparse shape-related local features for weak regressor training (Fig.~\ref{fig_2}).
Note that, for the first bounding box refinement step, we use only the dense local features.
\begin{figure}[t]
\centering
\subfloat[]{
 \label{fig_3_1}
 \includegraphics[trim = 4mm 156mm 130mm 0mm, clip, width = .7\linewidth]{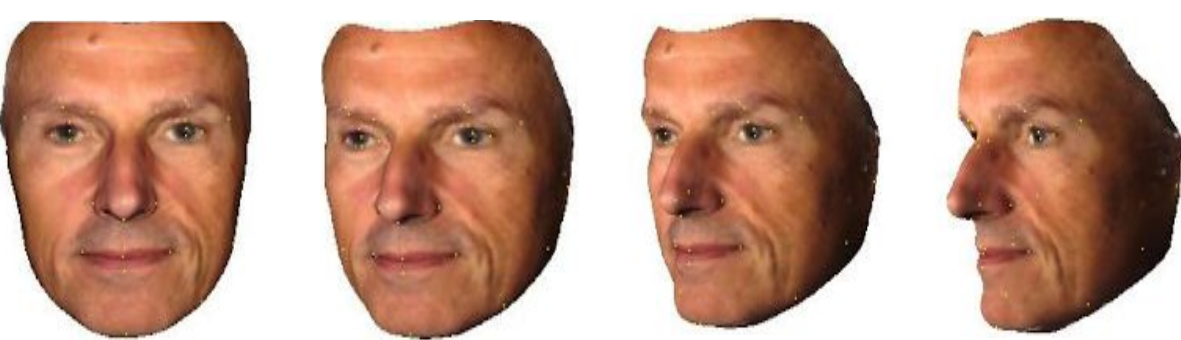}
}

\subfloat[]{
 \label{fig_3_2}
 \includegraphics[width=.2\linewidth]{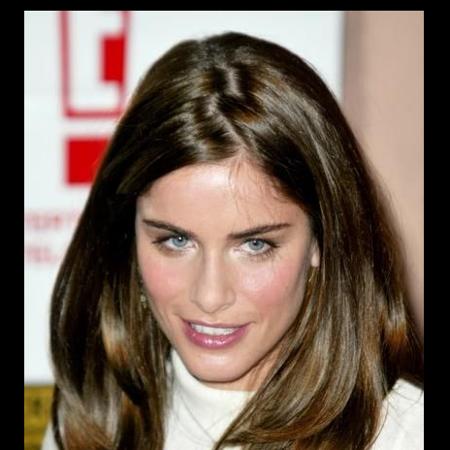}
 \includegraphics[width=.2\linewidth]{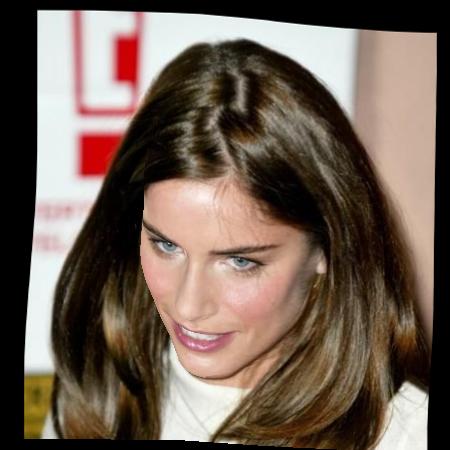}
 \includegraphics[width=.2\linewidth]{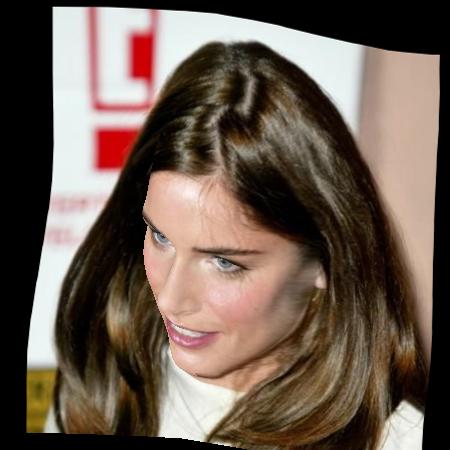}
 \includegraphics[width=.2\linewidth]{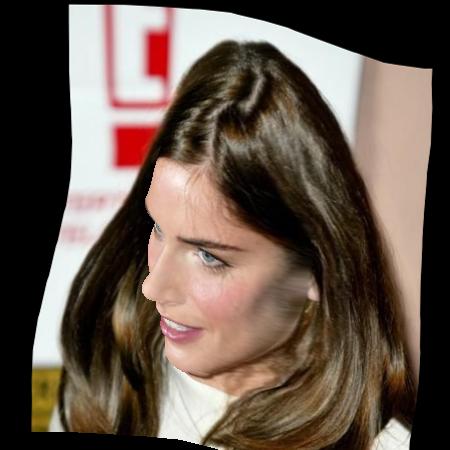}
}

\subfloat[]{
 \label{fig_3_3}
 \includegraphics[width=.2\linewidth]{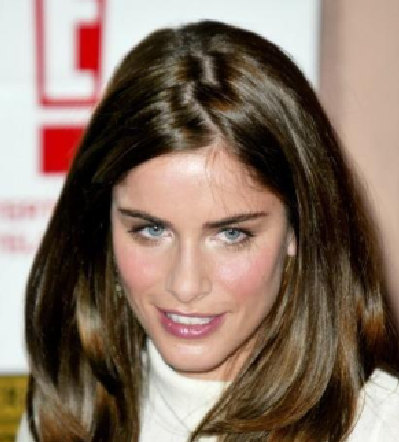}
 \includegraphics[width=.2\linewidth]{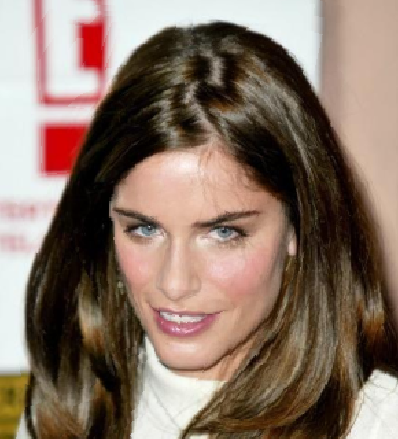}
 \includegraphics[width=.2\linewidth]{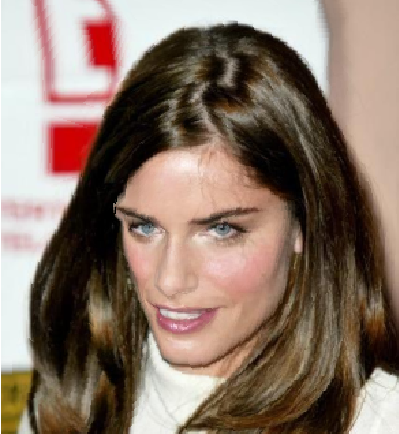}
 \includegraphics[width=.2\linewidth]{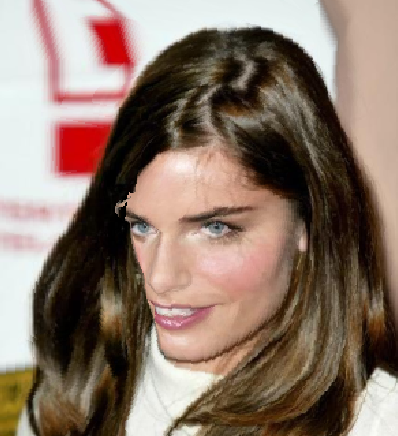}
}
\caption{A comparison of synthesised 2D faces using (a) a 3D morphable model~\cite{feng2015cascaded}, (b) 3D-based face profiling~\cite{zhu2016face}, and (c) our 2D-based method. }
\label{fig_3}
\end{figure}

\section{2D Profile Face Generation}
\label{sec_5}
For a learning-based approach, a large number of annotated face images are crucial for training.
As discussed in Section~\ref{sec_related_work}, traditional data augmentation methods are not able to inject new out-of-plane pose variations, and the use of 3D face models is very expensive.
To mitigate this issue, we propose a simple 2D-based method that can generate virtual faces with out-of-plane pose variations.
A comparison between our proposed 2D-based profile face generator and two 3D-based methods~\cite{feng2015cascaded,zhu2016face} is shown in Fig.~\ref{fig_3}.

\begin{figure}[t]
\centering
 \includegraphics[trim = 28mm 136mm 147mm 124mm, clip, width=.24\linewidth]{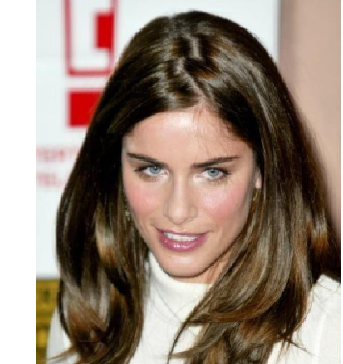}
 \includegraphics[trim = 110mm 136mm 65mm 124mm, clip, width=.24\linewidth]{./img/fig4.pdf}
 \includegraphics[trim = 70mm 136mm 105mm 124mm, clip, width=.24\linewidth]{./img/fig4.pdf}
 \includegraphics[trim = 153mm 136mm 22mm 124mm, clip, width=.24\linewidth]{./img/fig4.pdf}
\caption{The mesh generated for 2D image warpping.}
\label{fig_4}
\end{figure}

To warp a face image to another pose, we first build a PCA-based shape model that is equivalent to the shape model used in ASM~\cite{cootes1992active} and AAM~\cite{cootes1998active,Matthews2004}.
Then we choose the corresponding shape eigenvector controlling yaw rotations (usually the first one) to change the pose of the current face shape.
To this end, we first calculate the coefficient of the shape of a face image projected by the selected shape eigenvector.
A new face shape with pose variations is generated by adjusting the projected coefficient.
The 2D shape model used is constructed using a face dataset rich in pose variations.
Note, we only generate pose-varying face shapes with the same rotation direction of the original shape, \ie left or right.
Then we expand the face shape with more external facial landmarks and compute a 2D mesh of the original and new shapes using Delaunay triangulation, as shown in Fig.~\ref{fig_4}.
Last, a piece-wise affine warp is used to map the texture from the original face shape to a new one~\cite{Matthews2004}.
Moreover, the synthesised faces can be flipped about their vertical axis to obtain more faces with pose variations in the other direction (right or left), which is similar to~\cite{zhu2016face}.

\section{Experimental Results}
\label{sec_6}
\subsection{Datasets and Implementation Details}
\textbf{Datasets:}
In our experiments, we use two challenging face datasets, including the Annotated Facial Landmarks in the Wild (AFLW) dataset~\cite{aflw_dataset_2011} and the Caltech Occluded Faces in the Wild (COFW) dataset~\cite{Burgos-Artizzu2013}
to evaluate the performance of our DAC-CSR architecture.
\begin{table*}[t]
\renewcommand{\arraystretch}{1.2}
\centering
\caption{A summary of the evaluation protocols used in our experiments}
\small
\vspace{0.1cm}
\label{table1}
\begin{tabular}{lccccc}
\hline
Protocol & Training Set & Test Set & \# Landmarks & Normalisation & Setting  \\ 
\hline
AFLW-full    &     20000 from AFLW   & 4386 from AFLW &  19 &  face size   & CCL~\cite{zhu2016unconstrained}    \\ 
AFLW-frontal    &     20000 from AFLW  & 1165 from AFLW &  19 &  face size   & CCL~\cite{zhu2016unconstrained}     \\
COFW  & 1345 from COFW &  507  from COFW &  29  &  eye distance &  standard~\cite{Burgos-Artizzu2013}  \\ 
\hline
\end{tabular}
\end{table*}

The AFLW dataset has 25993 unconstrained faces with large-scale pose variations up to $\pm90^\circ$.
Each AFLW face image has up to 21 landmarks of visible facial parts.
AFLW does not have a standard protocol for FLD evaluation; hence we follow the protocol used in Cascaded Compositional Learning (CCL)~\cite{zhu2016unconstrained}.
This is the first work to use the whole AFLW dataset to benchmark an FLD algorithm.
It reports the currently best results on AFLW.
CCL used 24386 images from AFLW and manually annotated all the missing landmarks in the original dataset.
The annotation system opted for 19 landmarks per image without the two ear landmarks (ID-13 and ID-17).
CCL has two protocols: AFLW-full and AFLW-frontal, as shown in Table~\ref{table1}.
AFLW-full splits the 24386 images into 20000/4386 for training/testing.
The AFLW-frontal protocol selects 1165~\footnote{In our experiments, 1314 frontal faces were selected using the list provided by~\cite{zhu2016unconstrained}.} frontal images from the 4386 test images to evaluate an FLD algorithm on frontal faces. 

The COFW dataset has 1345 training and 507 test images, which are all unconstrained faces.
Each COFW face has 29 manually annotated landmarks.
COFW is a challenging benchmark containing major occlusions.

\textbf{Implementation Details:}
In our experiments, we only used one weak regressor for face bounding box refinement.
The numbers of weak regressors for general CSR and domain-specific CSR were set to 2 and 3 respectively.
We set the number of sub-domains to $M=5$ using 2 PCA shape coefficients, \ie $K=2$.
The value of the regularisation term in the ridge regression training was assigned to $\lambda = 10000$, and the decreasing schedule controlling fuzzy membership values was set to $h(n) = (0.3, 0.2, 0.1)$ for $n = (1, 2, 3)$.
To extract a dense face description, we resized the face region to $100\times 100$ and extracted HOG features using a cell size of 10 and block size of $2$.
To extract sparse shape-related local features, we computed the HOG descriptor in the neighbourhood of each facial landmark.
The radius was set to $1/7$ of the maximum of the height and width of the current shape estimate.
Each local image patch was resized to $30 \times 30$ and the cell size was set to 10.
In addition, the central $15 \times 15$ image patch was used to extract multi-scale HOG features using a cell size of 5.

To augment training data, we applied our 2D-based method to generate virtual face images with new poses.
Each training image in COFW was augmented using 9 new poses.
For AFLW, we only synthesised new faces for semi-frontal training images.
We also flipped all the training images about the vertical axis, added Gaussian blur with $\sigma = 1$ pixel and performed random perturbations of the initial face bounding boxes.
\begin{figure}[t]
\centering
 \includegraphics[trim = 38mm 95mm 40mm 100mm, clip, width=.9\linewidth]{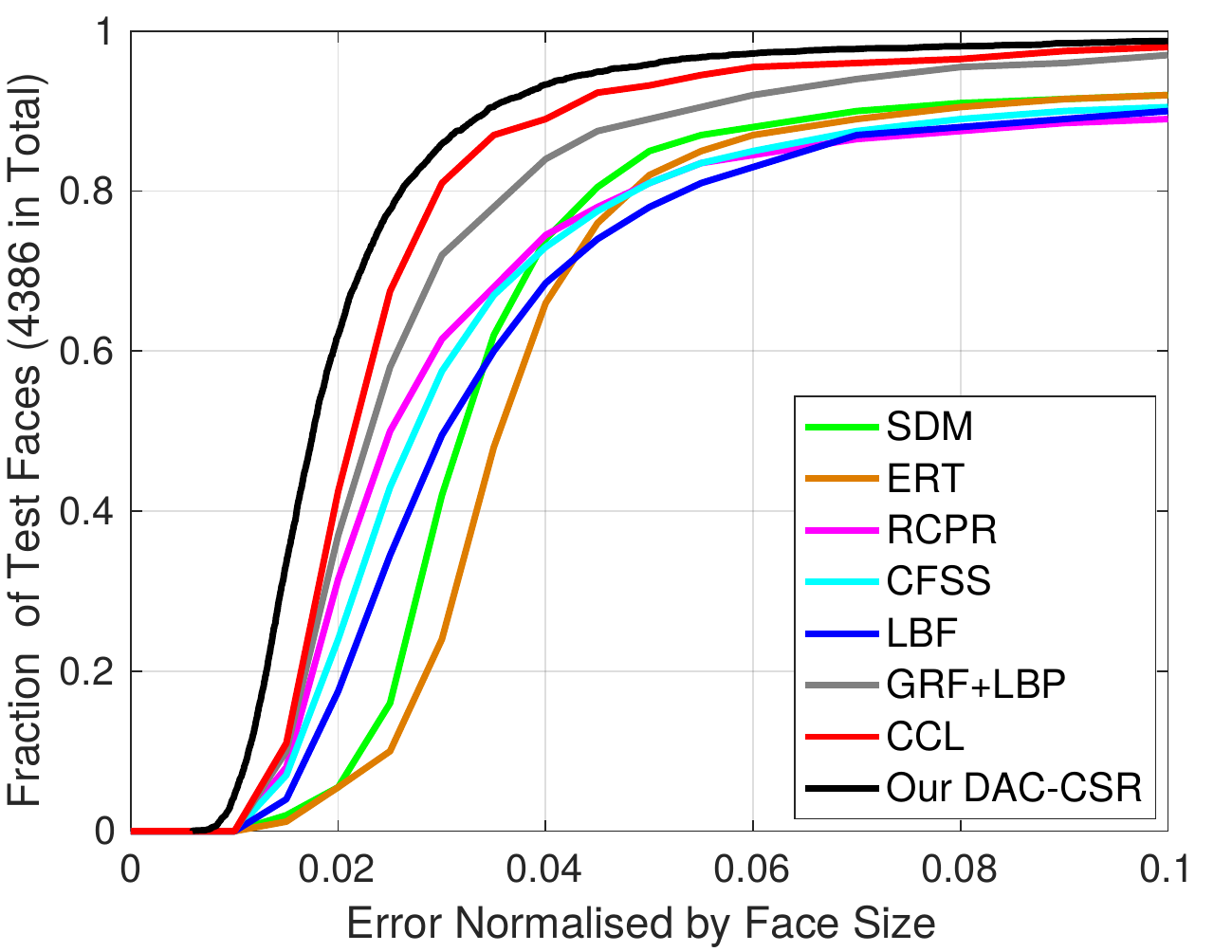}
\caption{A CED curve comparison of our DAC-CSR with state-of-the-art methods, including SDM~\cite{xiong2013supervised}, ERT~\cite{kazemi2014one}, RCPR~\cite{Burgos-Artizzu2013}, CFSS~\cite{zhu2015face}, LBF~\cite{ren2014face}, LBF + GRF~\cite{hara2014growing} and CCL~\cite{zhu2016unconstrained}, on the AFLW dataset (better viewed in colour). In this experiment, 20000 images were used for training and 4386 images were used for testing, following the \textbf{AFLW-full} protocol in~\cite{zhu2016unconstrained}.}
\label{fig_aflw_full}
\end{figure}
\begin{table}[t]
\renewcommand{\arraystretch}{1.1}
\centering
\caption{A comparison of our DAC-CSR with state-of-the-art methods on \textbf{AFLW}, measured in terms of the average error, normalised by face size. The protocol is the same as in~\cite{zhu2016unconstrained}.}
\label{table2}
\small
\vspace{0.1cm}
\begin{tabular}{lcc}
\hline
Method & AFLW-full & AFLW-frontal\\ 
\hline
SDM~\cite{xiong2013supervised} & 4.05\% & 2.94\% \\
RCPR~\cite{Burgos-Artizzu2013} & 3.73\% & 2.87\% \\
ERT~\cite{kazemi2014one} & 4.35\% & 2.75\% \\
LBF~\cite{ren2014face} & 4.25\% & 2.74\% \\
LBF + GRF~\cite{hara2014growing} & 3.15\% & N.A. \\
CFSS~\cite{zhu2015face} & 3.92\% & 2.68\% \\
CCL~\cite{zhu2016unconstrained} & 2.72\% & 2.17\% \\
\textbf{Our DAC-CSR} & \textbf{2.27\%} & \textbf{1.81{\Large }\%} \\
\hline
\end{tabular}
\end{table}

\subsection{Evaluation on AFLW}
The Cumulative Error Distribution (CED) curve of our DAC-CSR using the AFLW-full protocol is shown in Fig.~\ref{fig_aflw_full}.
The error was calculated using the Euclidean distance between the detected and ground-truth landmarks, normalised by face size~\cite{zhu2016unconstrained}.
Our DAC-CSR achieves much better results on the AFLW-full protocol than the current best result reported for CCL~\cite{zhu2016unconstrained}.

Table~\ref{table2} compares our DAC-CSR with state-of-the-art methods on AFLW using both the AFLW-full and AFLW-frontal protocols.
The results obtained with our DAC-CSR show the best normalised average error on both the full test set and the frontal face subset protocols.

\subsection{Evaluation on COFW}
\subsubsection{Comparison to State-of-the-art}
\begin{figure}[t]
\centering
 \includegraphics[trim = 38mm 95mm 40mm 100mm, clip, width=.9\linewidth]{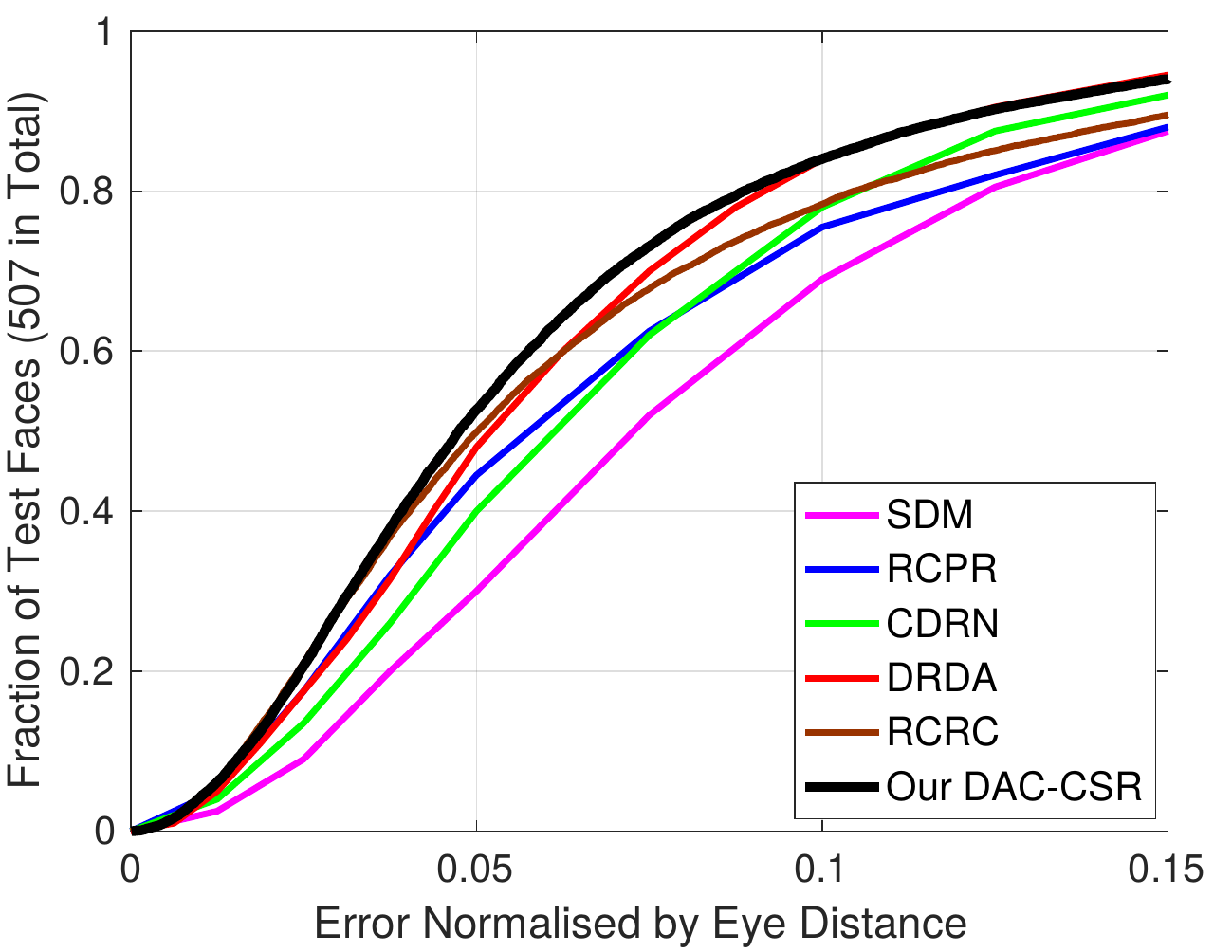}
\caption{A comparison between our DAC-CSR and state-of-the-art methods, including SDM~\cite{xiong2013supervised},  RCPR~\cite{Burgos-Artizzu2013}, RCRC~\cite{FENG2015}, CDRN~\cite{Zhang_2016_CVPR} and DRDA~\cite{Zhang_2016_CVPR}, on \textbf{COFW}.}
\label{fig_cofw}
\end{figure}

The CED curves of our DAC-CSR and a set of state-of-the-art methods on the COFW dataset are shown in Fig.~\ref{fig_cofw}. 
In addition, a more detailed comparison is presented in Table~\ref{table_3}, reporting the average error, failure rate and speed.
The failure rate is defined by the percentage of test images with more than $10\%$ detection error.
\begin{table}
\renewcommand{\arraystretch}{1.1}
\caption{Comparison on \textbf{COFW}. The error was measured on 29 landmarks and normalised by the inter-ocular distance.}
\label{table_3}
\centering
\small
\vspace{0.1cm}
\begin{tabular}{lccc}
\hline
               Method                      & Error & Failure  & Speed (FPS) \\
\hline
ESR~\cite{Cao2012}       & 11.2\%       	& 36\%            	& 4   \\
RCPR~\cite{Burgos-Artizzu2013}    & 8.5\%        	& 20\%            	& 3   \\
HPM~\cite{Ghiasi_2014_CVPR}     & 7.5\%        	& 13\%            	& 0.03   \\
RCRC~\cite{FENG2015}                 & 7.3\%        	& 12\%            	& 22  \\
CCR~\cite{feng2015cascaded}     & 7.03\%  & 10.9\% & \textbf{69} \\
DRDA~\cite{Zhang_2016_CVPR} & 6.46\% & 6\% & N.A. \\
RAR~\cite{Xiao2016} & 6.03\% & \textbf{4.14\%} & 4 (GPU)\\
\textbf{Our DAC-CSR} & \textbf{6.03\%} & 4.73\% & 10 \\
\hline
\end{tabular}
\end{table}

Our DAC-CSR achieves competitive results in accuracy compared to the two cutting-edge deep-neural-network-based algorithms, DRDA~\cite{Zhang_2016_CVPR} and RAR~\cite{Xiao2016}.
In addition, the speed of our DAC-CSR on an Intel i7-4790 CPU is up to 10 FPS, which is faster than RAR with GPU acceleration (NVIDIA Titan Z).
As the current bottleneck for unconstrained FLD is not the speed, \eg LBF can perform FLD at up to 3000 FPS, the key aim of our DAC-CSR is to provide a more robust FLD algorithm for faces with extreme appearance variations, as exhibited in the AFLW-full evaluation.

\subsubsection{Self Evaluation}
\label{sec_self_eva}
\begin{figure}[t]
\centering
 \includegraphics[trim = 38mm 95mm 40mm 100mm, clip, width=.9\linewidth]{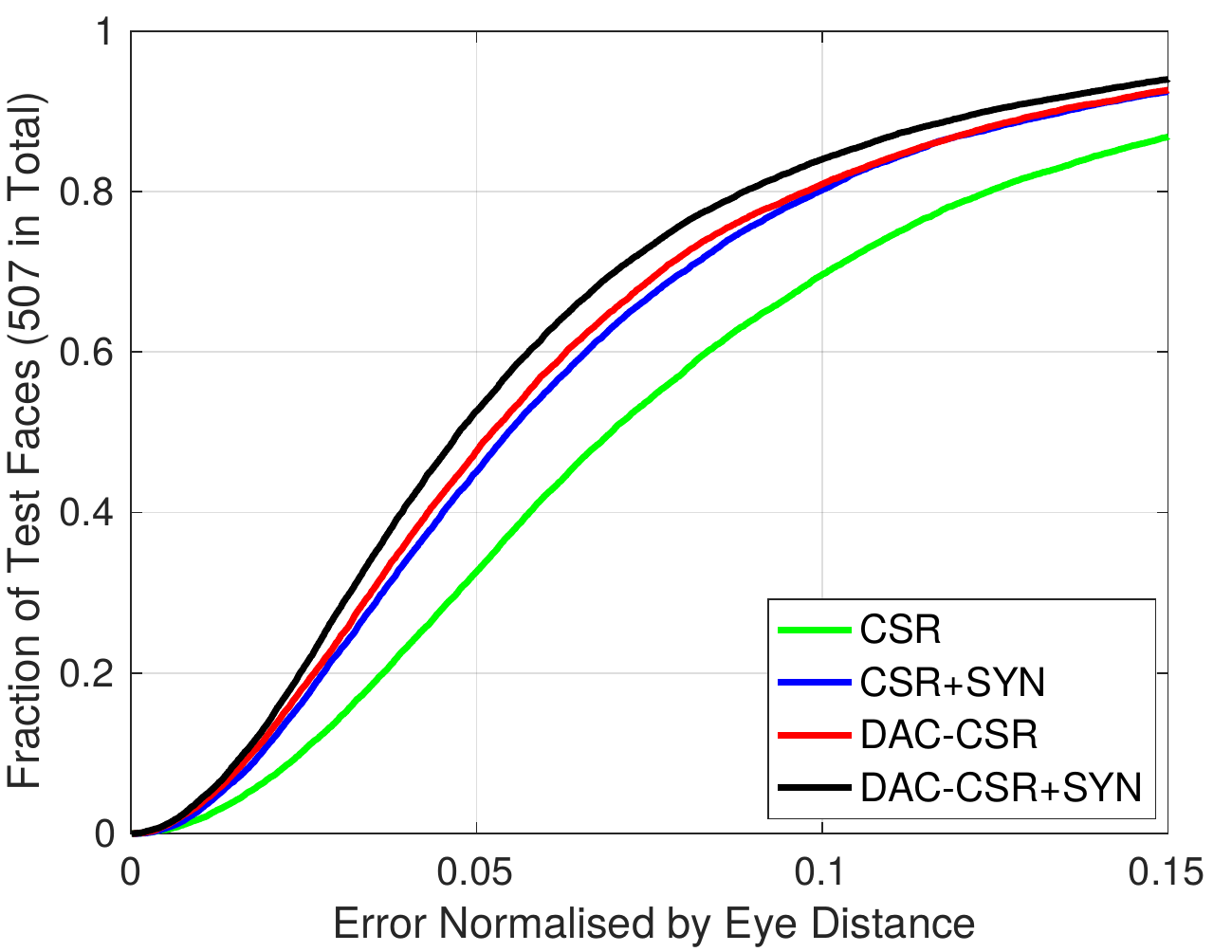}
\caption{A self-evaluation of our proposed DAC-CSR on \textbf{COFW}. The meaning of each term is introduced in Section~\ref{sec_self_eva}.}
\label{fig_cofw_2}
\end{figure}

In this part, we investigate the contributions of the proposed DAC-CSR architecture and our 2D-based data augmentation method to the accuracy of FLD on COFW. 
To this end, we compare the classical CSR method trained on the original training set (CSR) with the classical CSR trained on the augmented dataset using faces synthesised by our 2D-based face generation method (CSR+SYN), our DAC-CSR trained on the original dataset (DAC-CSR) and our DAC-CSR trained on the augmented dataset (DAC-CSR+SYN).
The CED curves of these settings are shown in Fig.~\ref{fig_cofw_2}.

In fact, the architecture of classical CSR is the same as SDM~\cite{xiong2013supervised}.
They also have similar CED curves (comparing Fig.~\ref{fig_cofw} with Fig.~\ref{fig_cofw_2}).
As indicated by Fig.~\ref{fig_cofw_2}, the new DAC-CSR architecture trained on the original dataset performs better than CSR with our 2D-based data augmentation method (DAC-CSR vs CSR+SYN).
However, the best result is achieved when the new DAC-CSR architecture is used jointly with our 2D-based data augmentation method.

\section{Conclusion}
\label{sec_7}
We have presented a new DAC-CSR architecture for robust FLD in unconstrained faces.
The proposed method achieved superior FLD results on the challenging AFLW dataset and delivered competitive performance on the COFW dataset. This is due to the proposed versatile fault-tolerant mechanism using fuzzy domain-specific model training and the online dynamic model selection strategy.
In addition, a simple but effective data augmentation method  based on 2D face synthesis was proposed. Compared with the classical CSR method, both the new DAC-CSR architecture and the 2D-based data augmentation method proved beneficial for the FLD performance on unconstrained faces.

We believe that our contributions can be further extended, \eg using deep-neural-network-based approaches.
We leave for future work the exploration of methods that combine our DAC-CSR architecture and data augmentation method with other FLD algorithms.

\section*{Acknowledgements}
This work was supported in part by the  EPSRC Programme Grant `FACER2VM' (EP/N007743/1), the National Natural Science Foundation of China (61373055, 61672265) and the Natural Science Foundation of Jiangsu Province (BK20140419, BK20161135).

{\small
\bibliographystyle{ieee}
\bibliography{mybib}
}

\end{document}